\documentclass[letterpaper]{article} 
\usepackage{aaai2027}  
\usepackage[hyphens]{url}  
\usepackage{graphicx} 
\usepackage{natbib}  
\usepackage{caption} 
\usepackage{amsmath}
\usepackage{booktabs}
\usepackage{multirow}
\usepackage{array}
\newcolumntype{R}[1]{>{\raggedright\arraybackslash\hyphenpenalty=10000%
  \let\\\tabularnewline}p{#1}}
  
\newcommand{\sys}{TRACE}

\title{Auditable Emergency Triage for Maternal and Newborn Care in India}
\author{
    Shobhit Jagga\equalcontrib\textsuperscript{\rm 1},
    Aman Dalmia\equalcontrib\textsuperscript{\rm 2},
    Niharika Priyadarshini\equalcontrib\textsuperscript{\rm 1},
    Neelima Devadas\textsuperscript{\rm 1},
    Amrita K Prasen\textsuperscript{\rm 1},\\
    Nikhil Nalin\textsuperscript{\rm 1},
    Santhosh SJ\textsuperscript{\rm 1},
    Sreeram Nurani Ramasubramanian\textsuperscript{\rm 1},
    Muhammed Afeer K\textsuperscript{\rm 1},
    Anubhav Arora\textsuperscript{\rm 1}
}
\affiliations{
    \textsuperscript{\rm 1}Noora Health\\
    \textsuperscript{\rm 2}The Agency Fund\\
    shobhit@noorahealth.org, aman@agency.fund, niharika@noorahealth.org, neelima@noorahealth.org, amrita@noorahealth.org, nikhil@noorahealth.org, santhosh@noorahealth.org, sreeram@noorahealth.org, afeer@noorahealth.org, anubhav@noorahealth.org
}

\begin{document}
\maketitle

\begin{abstract}
At Noora Health, our nurses answer more than 50,000 medical queries per month on our WhatsApp-based service that provides caregivers with on-demand support. Their most time-critical task is emergency triage: deciding which queries need immediate in-person attention. To support them, we built a system that uses a large language model (LLM) to classify whether a message is an emergency and provide a rationale for interpretability. Nurses can flag whether the system missed an emergency or incorrectly labeled it as one, giving us a live measure of missed emergencies and false alarms. But the system was opaque: analyzing mistakes meant reading reasoning chains for each message, which is infeasible at our scale. Prompt changes meant re-running a full evaluation to prevent regressions, which was both costly and operationally challenging. Clinicians follow a decision tree to make this call, but it was never documented or passed to the model, which relied on a flat list of danger signs. To address these issues, we decomposed triage into two steps: an LLM extracts canonical symptoms and patient context from the query using a clinician-authored vocabulary, and a deterministic rule engine captures the scenarios that indicate an emergency. We show that the new system raised recall from 0.565 to 0.810 and F1 from 0.606 to 0.702, with structured rules driving most of the accuracy gains while the decomposition provides auditability: clinical experts can inspect each stage of the new system to see whether the query was mistranslated, symptoms were incorrectly extracted, patient context was wrongly inferred, or the necessary rules were missing. They can add new rules independently without causing regressions and avoid running costly evaluations. Since deployment, the new system has triaged 152,421 patient queries and flagged 28,535 (18.7\%) as emergencies. The over-escalation rate has been 17.8\%, without any increase in missed emergencies. Clinicians have also added 48 new rules since deployment, evidence of the faster correction loop we set out to build.
\end{abstract}

\section{Introduction}

Noora Health, a non-profit organization, supports the design of the Care Companion Program (CCP), which trains families on essential maternal and neonatal care during their hospital stay and is implemented with local partners across district hospitals in nine states of India. After discharge, families receive follow-up through a dedicated WhatsApp-based platform that allows two-way communication: they can ask questions and receive guidance from our support staff, both doctors and nurses. This platform receives more than 50,000 medical queries a month, mostly spanning maternal and newborn care, alongside other conditions that CCP covers, such as cardiac and chronic disease care. Our nurses answer routine questions directly and refer emergencies for immediate in-person care. But identifying them in a timely and reliable manner is hard.

\begin{figure*}[t]
  \centering
  \includegraphics[width=\textwidth]{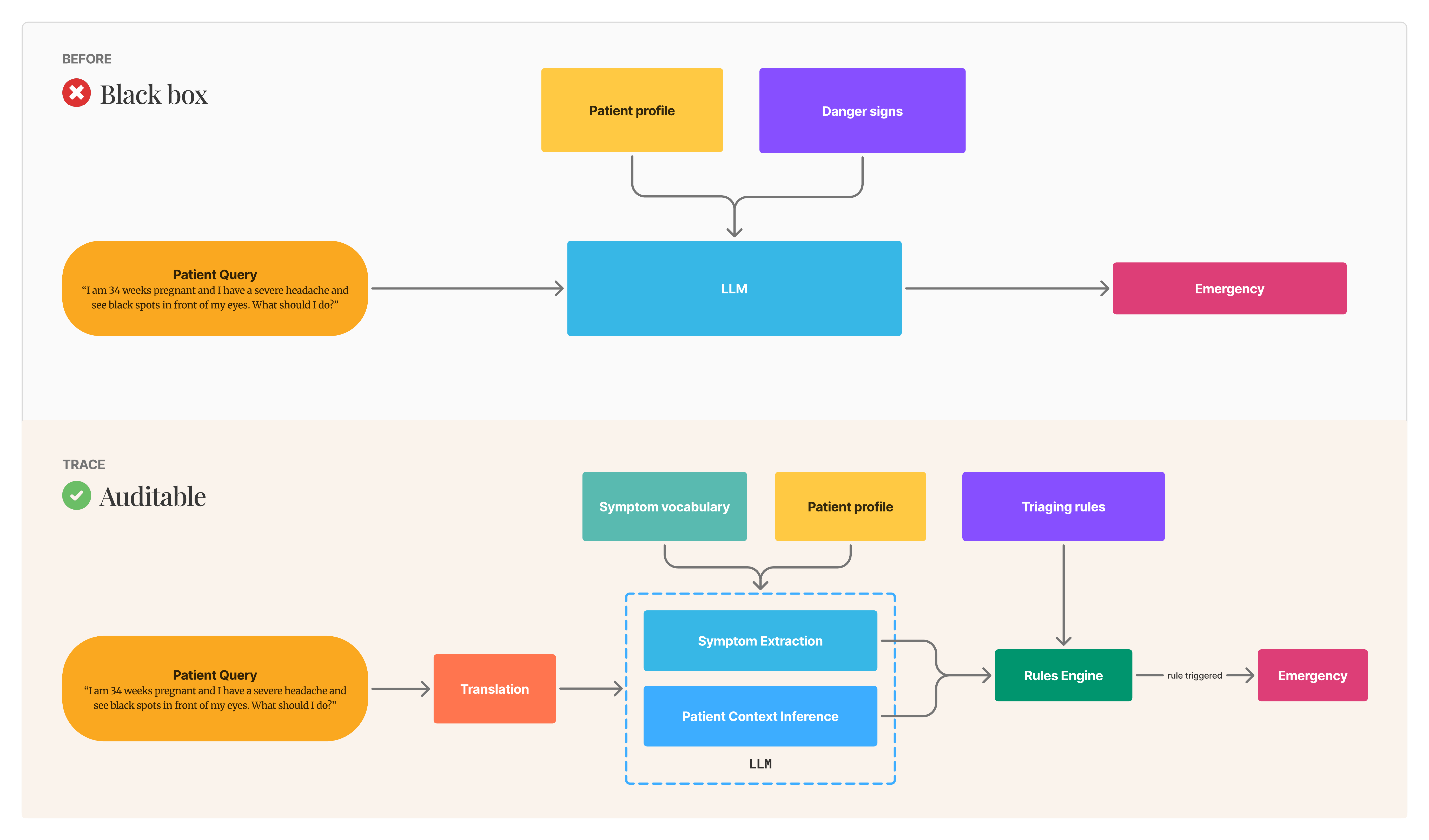}
  \caption{Evolution of our AI-assisted emergency triage system. The initial version (top) uses an LLM to predict, end-to-end, whether the query indicates an emergency, with a list of danger signs as the knowledge base. \sys{} (bottom) decomposes the task into two steps: LLM-based symptom extraction and patient context inference, followed by a deterministic rule engine, so every decision can be traced to the specific rule that drove it.}
  \label{fig:branch}
\end{figure*}

LLMs are being increasingly deployed in healthcare~\citep{thirunavukarasu2023}, including in low-resource, multilingual settings in low- and middle-income countries (LMICs), such as ours, where demand for clinical guidance far exceeds the supply of trained providers~\citep{chen2025lmics}. A missed escalation can directly delay life-saving care. Our messages make the task harder: they are short, code-switched across regional languages~\citep{khullar-etal-2025-script,agarwal2025}, and describe symptoms in colloquial terms rather than standardised clinical vocabulary~\citep{zeng2006}. They often omit context, yet the same symptom can signal a different level of urgency depending on the patient context: a fever in a newborn is a medical emergency, while the same fever in an older child usually is not. The stakes are high: India recorded an estimated 19,000 maternal deaths in 2023, the second-highest number of any country~\citep{who2025}, and accounts for roughly 18\% of neonatal deaths globally, one of the largest shares in the world~\citep{unicef2024}. Delays in recognizing danger signs are among the leading preventable causes.

We initially built our AI-assisted triage system by prompting an LLM, along with a list of danger signs commonly associated with an emergency, to classify the message and generate a reasoning chain for its decision (Figure~\ref{fig:branch}). It gave promising results on our initial test set and was subsequently deployed. Our nurses can see which messages have been flagged as an emergency and override an incorrect prediction, giving us a continuous measure of over-escalations and missed emergencies in deployment. Since missed emergencies are more harmful, the system was optimized for higher recall. It caught real emergencies faster, but the number of over-escalations at our scale became a challenge.

The system was also opaque. The only way to analyse mistakes was by reading the reasoning generated by the LLM, one message at a time, which is not practical at our scale. It did not match how our clinicians work either: they follow a decision tree that factors in patient context and symptom combinations, learned through their training and experience rather than documented for a model to use. Our prompt carried only a flat list of danger signs, so any change risked degrading performance. To prevent regressions, we had to run a full evaluation before any change was deployed, which increased our costs, required different teams to coordinate, and increased the time taken to resolve issues. These problems share a common root cause: the model was doing two jobs at once, interpreting the patient query and making the clinical decision.

To address this, we built \sys{} (\textbf{T}riage by \textbf{R}ule-based \textbf{A}pplication on \textbf{C}ontext-aware \textbf{E}xtractions). It separates the two jobs as a clinician would: an LLM extracts canonical symptoms from a fixed, clinician-curated vocabulary along with patient context such as trimester or infant age, and a deterministic rule engine, owned by the clinical team, decides whether to mark it an emergency. Symptom extraction is a narrower task than triage and is the only step that needs subjective interpretation of the query. Once the symptoms and context are extracted, the rules are fixed. Figure~\ref{fig:branch} compares the original end-to-end LLM system with \sys{}.

We show that structured rules drive most of the accuracy gains, raising recall from 0.565 to 0.810 and F1 from 0.606 to 0.702 over the system \sys{} replaced, while the decomposition provides auditability as every mistake is traceable to a specific step: mistranslation of the query, a symptom either not extracted or missing from the vocabulary altogether, the wrong patient context inferred, or a rule that was wrong or missing. Each category has an owner: missing rules or symptoms are handled by clinicians and extraction errors are analysed jointly with the Machine Learning (ML) team. Since deployment, it has triaged 152,421 patient queries and flagged 28,535 (18.7\%) as emergencies, with an over-escalation rate of 17.8\% and no increase in missed emergencies. Clinicians add new rules (48 so far) without touching the LLM prompt, so a change no longer requires re-running an expensive evaluation.

\section{Related Work}

\paragraph{Clinical triaging with NLP and LLMs.} Recent work shows that LLMs can read unstructured presentations and rate acuity near clinician level, on both emergency-department notes~\citep{williams2024} and curated vignettes, where the typical failure is over-estimating urgency~\citep{sorich2024}. These results assume clean clinical inputs, and the systems built on them keep the verdict inside the model, reached by multi-agent deliberation or retrieval over a triage handbook~\citep{han2024}. The settings closest to ours are deployed maternal and child health support services in low-resource regions: TRIM-AI triages code-mixed SMS in Kenya~\citep{zhang2023}, a WhatsApp chatbot performs stage-aware maternal triage in India~\citep{jha2026}, and CLARITY pairs a deterministic state machine with LLM agents to route patients to specialists on a national platform~\citep{shaposhnikov2025}. In each, either the emergency decision lives inside a generative model or the deterministic component governs the conversation rather than the verdict. Our system instead confines the LLM to symptom extraction and hands the emergency decision to a clinician-editable rule engine, a separation no prior triage system in this setting makes.

\paragraph{Multilingual NLP in low-resource healthcare.} Patients rarely describe symptoms in clinical vocabulary, a gap that has motivated consumer-health vocabularies for two decades~\citep{zeng2006}; we follow this with a clinician-curated mapping from lay phrases to canonical symptom keywords. The messages are short and code-switched across regional languages, often in romanized script, which remains hard for NLP~\citep{khanuja2020}. Existing LLMs degrade on these languages~\citep{singh2024,agarwal2025}. \citet{khullar-etal-2025-script} shows that on real maternal-triage queries, romanization costs up to 24 F1 points even when the model infers the correct intent, placing the failure in the final classification rather than in comprehension. This is precisely the brittle step our architecture moves out of the LLM and into a deterministic rule layer.

\paragraph{Interpretability and deployment.} For high-stakes decisions, a long line of work argues that systems should be inherently interpretable rather than explained after the fact~\citep{rudin2019}, since post-hoc explanations are unreliable for individual clinical cases~\citep{ghassemi2021}. A reviewable basis for each recommendation is also the regulatory line between decision support and a regulated device~\citep{fda2022}. The standard remedy treats the LLM as a language front-end and delegates the decision to a deterministic component, which, at the physician level, yields fully traceable labels when LLM extraction is paired with a rule-based expert system~\citep{prenosil2025}. The closest analogue is DORIS, which annotates text against clinical criteria and trains a deterministic classifier on the result~\citep{lan2025}; we share the decomposition but use clinician-editable rules that change without retraining, and run the LLM at inference time on live messages. Deployment studies reinforce the value of this lever: expert corrections improved CataractBot's accuracy and cut workload~\citep{sachdeva2024}, while ASHABot's users treat its output as authoritative, an argument for hard guardrails~\citep{ramjee2025}. Every verdict in our system traces to a specific rule a clinician can audit and modify directly.

\section{System Design}

\subsection{Problem Formulation}

The system triages one message at a time. We formalize the input as a tuple $(m,\,c,\,P)$. Here, $m$ is the patient message. $c$ represents the care category under which the patient is enrolled with us, one of the eight we support: antenatal care (ANC), postnatal care (PNC), special newborn care unit (SNCU), high-risk pregnant women (HRPW), general health and wellness (GHW), non-communicable disease (NCD), maternal and child health (MCH), and general medicine and surgery (GEMS) (including cardiac and post-surgery care). $P$ is the profile information we maintain for every patient. For a pregnant patient, it may contain the expected delivery date and the gestational trimester $t \in \{1, 2, 3\}$ she is currently in. For a patient who has already given birth, it stores the date of delivery and the infant age instead. Triage cannot be done from the message alone: the urgency of the same symptom varies with the patient context. The output of the system is $y \in \{\text{emergency},\,\text{non-emergency}\}$.

\begin{table}[t]
\centering
\small
\begin{tabular}{ll}
\toprule
\textbf{How patients say it} & \textbf{Symptom it maps to} \\
\midrule
bleeding from vagina        & Vaginal bleeding \\
bleeding ho rahi hai        & Vaginal bleeding \\
passing large clots         & Vaginal bleeding \\
baby is not moving          & Reduced fetal movement \\
bachcha move nahi kar raha  & Reduced fetal movement \\
fits / jhatkay              & Seizure \\
chest mein dard             & Chest pain \\
\bottomrule
\end{tabular}
\caption{Illustrative vocabulary entries. Patients describe the same symptom in English, in romanized Hindi, and code-mixed within a single phrase.}
\label{tab:vocab}
\end{table}
\begin{table*}[t]
\centering
\small
\setlength{\tabcolsep}{4pt}
\begin{tabular}{llll}
\toprule
\textbf{Care category} & \textbf{Symptom} & \textbf{Also required} & \textbf{Emergency?} \\
\midrule
\multicolumn{4}{@{}l}{\textit{Standalone}} \\
\quad GENERAL & Seizure                & nothing                              & Yes \\
\quad HRPW    & Fever above 38.5$^\circ$C for 24 hours & nothing              & Yes \\
\midrule
\multicolumn{4}{@{}l}{\textit{Combination: symptom + symptom}} \\
\quad ANC     & Burning urination      & no fever                             & No \\
\quad ANC     & Burning urination      & with fever                           & Yes \\
\quad ANC     & Leg pain               & with leg swelling and redness        & Yes \\
\quad PNC     & Leg swelling           & with chest pain or breathing trouble & Yes \\
\midrule
\multicolumn{4}{@{}l}{\textit{Combination: symptom + context}} \\
\quad ANC     & Reduced fetal movement & trimester 1                          & No \\
\quad ANC     & Reduced fetal movement & trimester 2--3                       & Yes \\
\bottomrule
\end{tabular}
\caption{Example rules. Rules are scoped to a care category, so the same symptom is mapped to a different rule depending on what the patient is enrolled with us for: leg pain while she is pregnant (ANC) needs swelling and redness to be considered an emergency, while leg swelling post-delivery (PNC) becomes an emergency only if she has chest pain or breathing trouble. A \textit{Standalone} rule applies on the symptom alone. A \textit{Combination} rule applies only when a second symptom is present, or when the patient context matches, such as a given trimester or infant age. HRPW is a pregnancy already flagged as high-risk, and GENERAL rules apply to any patient whatever their care category.}
\label{tab:rules}
\end{table*}

\subsection{Two-Step Decomposition}
\label{sec:sys-overview}

Queries span seven languages and are usually typed in Roman script rather than in their native script, which degrades quality on this task~\citep{khullar-etal-2025-script}. So, every query is first translated into English using Gemini-2.5-Flash. An LLM then extracts the canonical symptoms the message describes, along with any patient context it contains, and a short rationale. The rule engine then evaluates those symptoms and context together with $c$ and $P$ to assign the emergency (or not) label. The LLM never makes the final emergency decision.

\section{Symptom Vocabulary and Structured Rules}
\label{sec:knowledge}

\paragraph{Symptom vocabulary.} A patient may express the same symptom in multiple ways, often using colloquial terms instead of the standard clinical terminology. Our clinicians created a symptom vocabulary that maps the different ways a patient might phrase a symptom to its canonical keyword, following the long-standing practice of bridging everyday language to clinical concepts through a consumer-health vocabulary~\citep{zeng2006}. This is given as a reference to the LLM which extracts all the canonical symptoms present in the user query. Table~\ref{tab:vocab} shows a few examples. The vocabulary map covers \textbf{745} patient phrasings mapped to \textbf{184} canonical symptoms across the eight care categories.

\paragraph{Structured rule engine.} The rules define when the presence of one or more canonical symptoms constitutes an emergency for the given patient context. A \textit{Standalone} rule applies whenever the corresponding symptoms are found, whatever the patient context (e.g.\ convulsions, severe vaginal bleeding, difficulty breathing), whereas a \textit{Combination} rule applies only when a symptom appears together with a second symptom or a specific patient context: abdominal pain by itself is not an emergency, but if it occurs along with vaginal bleeding, it should be flagged. Table~\ref{tab:rules} shows a few illustrative examples. The clinical team prepared 231 rules in total: 141 \textit{Standalone} and 90 \textit{Combination}. 

\paragraph{Generalization of the decision tree.} Neither the symptom vocabulary nor the rule engine is meant to enumerate every possible emergency, but only those our program supports. The clinical team developed these through extensive analysis of production queries before the dataset in Section~\ref{sec:dataset} was built. For queries that the current rules or symptom vocabulary do not cover, clinicians revise the rule set or the vocabulary map accordingly.

\section{Dataset}
\label{sec:dataset}

\begin{table}[t]
  \centering
  \small
  \begin{tabular}{lrr}
    \toprule
    Language & N & \% \\
    \midrule
    Hindi   & 255 & 33.2 \\
    English & 177 & 23.0 \\
    Telugu  & 152 & 19.8 \\
    Kannada &  65 &  8.5 \\
    Punjabi &  52 &  6.8 \\
    Marathi &  46 &  6.0 \\
    Odia    &  22 &  2.9 \\
    \midrule
    Total   & 769 & 100 \\
    \bottomrule
  \end{tabular}
  \caption{Distribution of the dataset by language}  \label{tab:dist-language}
\end{table}

\begin{table}[t]
  \centering
  \small
  \begin{tabular}{lrr}
    \toprule
    Care category & N & \% \\
    \midrule
    ANC  & 377 & 49.0 \\
    PNC  & 175 & 22.8 \\
    HRPW & 130 & 16.9 \\
    SNCU &  28 &  3.6 \\
    GHW  &  20 &  2.6 \\
    MCH  &  15 &  2.0 \\
    GEMS &  13 &  1.7 \\
    NCD  &  11 &  1.4 \\
    \midrule
    Total & 769 & 100 \\
    \bottomrule
  \end{tabular}
  \caption{Distribution of the dataset by care category}
  \label{tab:dist-category}
\end{table}

Our dataset consists of real queries sent by patients to our WhatsApp chatbot. It was collected and annotated in two batches, giving a total of 769 queries containing 251 emergencies, across all the languages we support: Hindi, English, Telugu, Kannada, Punjabi, Marathi, and Odia. The distribution across languages and care categories is given in Tables~\ref{tab:dist-language} and~\ref{tab:dist-category}. The dataset was divided into 342 validation queries (104 emergencies) and 427 unseen test queries (147 emergencies). The first batch of data was stratified by care category to prepare the validation and test sets. The second batch of data was added entirely to the test set.

Two in-house clinicians independently labelled every query. They each had access to the same patient context that was made available to the triage system. Across the full dataset, the two annotators reached 83\% raw agreement (Cohen's $\kappa$ = 0.626) and disagreements were resolved by treating the label given by the senior clinician as the ground truth.

\begin{table*}[t]
\centering
\small
\begin{tabular*}{\textwidth}{@{\extracolsep{\fill}}lcccrr@{}}
\toprule
System & P & R & F1 & Latency (p50, secs) & Cost (cents/query) \\
\midrule
LLM-KB        & \textbf{.654} [.586, .724] & .565 [.483, .646] & .606 [.539, .669] & 4.49 & 0.099 \\
LLM-Rules     & .586 [.541, .633] & \textbf{.837} [.776, .891] & .689 [.645, .732] & 2.72 & \textbf{0.060} \\
\sys{}-NoCtx  & .626 [.577, .679] & .810 [.741, .871] & \textbf{.706} [.659, .752] & \textbf{1.96} & 0.082 \\
\sys{}$^\dagger$ & .620 [.571, .672] & .810 [.741, .871] & .702 [.655, .748] & 2.55 & 0.093 \\
\bottomrule
\end{tabular*}
\caption{Comparison of the architectures on the held-out test set ($N=427$, 147 emergencies) with Gemini-2.5-Flash as the LLM. Brackets show 95\% stratified-bootstrap confidence intervals. The translation step, shared by all systems, is excluded from latency and cost numbers.}
\label{tab:architecture}
\end{table*}
\begin{table*}[t]
\centering
\small
\setlength{\tabcolsep}{3pt}
\begin{tabular}{lrr|ccc|ccc}
\toprule
Language & N & Em. & \multicolumn{3}{c|}{LLM-Rules} & \multicolumn{3}{c}{\sys{}} \\
 & & & P & R & F1 & P & R & F1 \\
\midrule
Hindi & 127 & 45 & 0.61 [0.53, 0.70] & \textbf{0.87 [0.76, 0.96]} & \textbf{0.72 [0.64, 0.79]} & \textbf{0.64 [0.55, 0.74]} & 0.82 [0.71, 0.93] & \textbf{0.72 [0.64, 0.80]} \\
Telugu & 82 & 23 & 0.49 [0.40, 0.61] & 0.83 [0.65, 0.96] & 0.61 [0.51, 0.72] & \textbf{0.53 [0.43, 0.65]} & \textbf{0.87 [0.74, 1.00]} & \textbf{0.66 [0.55, 0.76]} \\
English & 76 & 31 & 0.68 [0.58, 0.81] & \textbf{0.84 [0.71, 0.97]} & 0.75 [0.66, 0.85] & \textbf{0.75 [0.63, 0.88]} & 0.77 [0.61, 0.90] & \textbf{0.76 [0.65, 0.87]} \\
Kannada & 40 & 13 & 0.53 [0.38, 0.71] & 0.69 [0.46, 0.92] & 0.60 [0.40, 0.77] & \textbf{0.56 [0.41, 0.75]} & \textbf{0.77 [0.54, 1.00]} & \textbf{0.65 [0.47, 0.80]} \\
Marathi & 40 & 16 & 0.57 [0.44, 0.71] & \textbf{0.81 [0.62, 1.00]} & \textbf{0.67 [0.53, 0.79]} & \textbf{0.60 [0.45, 0.78]} & 0.75 [0.50, 0.94] & \textbf{0.67 [0.50, 0.81]} \\
Punjabi & 40 & 11 & \textbf{0.50 [0.36, 0.69]} & \textbf{0.82 [0.55, 1.00]} & \textbf{0.62 [0.45, 0.79]} & \textbf{0.50 [0.33, 0.71]} & 0.73 [0.45, 1.00] & 0.59 [0.40, 0.77] \\
\bottomrule
\end{tabular}
\caption{Performance of LLM-Rules and \sys{} by language on the test set. \textit{Em.}\ is the number of emergencies. Odia is omitted, with only 22 queries and 8 emergencies.}
\label{tab:by_language}
\end{table*}

The size of the dataset is limited by the bandwidth of our clinical experts for review. What counts as an emergency is set by our own evolving clinical protocol, so the labels have to come from our in-house clinicians. Nurses do label every production query as they work, but under time pressure they can make mistakes or not follow the protocol perfectly when correcting an emergency prediction. So, we treat the labels given by clinicians as the definitive labels. That adjudication is additional work on top of clinical duties that already leave them with little spare time. Our government partnership agreements forbid sending patient data to commercial annotation vendors.

\begin{table*}[t]
\centering
\small
\setlength{\tabcolsep}{3pt}
\begin{tabular}{lrr|ccc|ccc}
\toprule
Care category & N & Em. & \multicolumn{3}{c|}{LLM-Rules} & \multicolumn{3}{c}{\sys{}} \\
 & & & P & R & F1 & P & R & F1 \\
\midrule
ANC & 177 & 54 & 0.64 [0.56, 0.73] & \textbf{0.85 [0.76, 0.94]} & \textbf{0.73 [0.66, 0.80]} & \textbf{0.65 [0.56, 0.75]} & 0.76 [0.65, 0.87] & 0.70 [0.61, 0.79] \\
PNC & 136 & 38 & 0.48 [0.41, 0.55] & \textbf{0.87 [0.76, 0.97]} & 0.62 [0.54, 0.69] & \textbf{0.51 [0.44, 0.59]} & \textbf{0.87 [0.76, 0.97]} & \textbf{0.64 [0.56, 0.72]} \\
HRPW & 69 & 32 & 0.59 [0.51, 0.69] & 0.81 [0.66, 0.94] & 0.68 [0.59, 0.78] & \textbf{0.64 [0.57, 0.74]} & \textbf{0.91 [0.81, 1.00]} & \textbf{0.75 [0.68, 0.83]} \\
\bottomrule
\end{tabular}
\caption{Performance of LLM-Rules and \sys{} by care category on the test set. \textit{Em.}\ is the number of emergencies. The categories with fewer than 40 queries have been omitted.}
\label{tab:by_category}
\end{table*}

\section{Experiments}
\label{sec:results}

\subsection{Setup}

We evaluate four systems as a stepwise ablation that isolates the contribution of each design choice:

\begin{itemize}
  \item \textbf{LLM-KB} (baseline): An end-to-end LLM system that generates the emergency label from a list of danger signs, along with a rationale. This is the initial production system that TRACE replaces.
  \item \textbf{LLM-Rules}: Same as LLM-KB with the list of danger signs replaced by the structured rules (Section~\ref{sec:knowledge}).
  \item \textbf{\sys{}-NoCtx}: The TRACE architecture without the patient context inference. The LLM extracts only the symptoms from the query.
  \item \textbf{\sys{}}: The full TRACE architecture (Section~\ref{sec:sys-overview}).
\end{itemize}

The validation set was used to iterate on the prompts, and the final metrics are reported on the test queries (Table~\ref{tab:architecture}). Recall is the primary safety metric because a missed emergency is more harmful than a false alarm. F1 is the secondary metric to balance recall with false alarms.

The same LLM, Gemini-2.5-Flash, and the same translated queries were used for every experiment. Since LLMs are stochastic, their outputs can vary on each run. To verify that our results are not driven by that variation, we ran the same configuration for each architecture three times on the validation set before running them on the test set. F1 ranged from 0.764 to 0.769 (mean 0.767, SD 0.002) and recall from 0.861 to 0.870 (mean 0.867, SD 0.005). The variation across runs is much smaller than the differences we observe across architectures. On the test set, we also report 95\% confidence intervals, calculated from 10,000 bootstrap resamples stratified by class.

\subsection{Structured rules drive the accuracy gain}
Replacing the list of danger signs (LLM-KB) with structured rules (LLM-Rules) alone raises recall from 0.565 to 0.837 and F1 from 0.606 to 0.689. The rules make the decision tree our clinicians follow explicit. Without them, the model falls back on its own reasoning, shaped by its training data rather than our protocols.

\subsection{Auditability has no measurable quality loss}

\sys{} reaches a recall of 0.810 [0.741, 0.871] and F1 0.702 [0.655, 0.748], against 0.837 [0.776, 0.891] and 0.689 [0.645, 0.732] for LLM-Rules. Since the confidence intervals overlap, we cannot say one architecture is more accurate on this dataset, with \sys{} providing the additional benefit of auditability as described in Section~\ref{sec:extraction-audit}. TRACE is slightly faster, with a median latency of 2.55 seconds per query, than LLM-Rules (2.72 seconds/query).

\begin{table*}[t]
\small
\setlength{\tabcolsep}{4pt}
\begin{tabular}{%
  >{\raggedright\arraybackslash}p{0.165\textwidth}
  >{\raggedright\arraybackslash}p{0.12\textwidth}
  >{\raggedright\arraybackslash}p{0.14\textwidth}
  >{\raggedright\arraybackslash}p{0.13\textwidth}
  >{\raggedright\arraybackslash}p{0.075\textwidth}
  >{\raggedright\arraybackslash}p{0.18\textwidth}
  >{\raggedright\arraybackslash}p{0.075\textwidth}}
\toprule
Query & Patient profile & LLM extraction & Rules activated & Error & Cause and fix & Owner \\
\midrule
Pregnant woman's blood pressure has increased, BP is 167 & GHW, no due date & severe hypertension; context: ANC & none; rule is HRPW-only & missed alarm & rule too narrow $\rightarrow$ widen to all pregnancy & Clinicians \\
\addlinespace
The operation date has passed and 9 months have been completed & ANC, due 2 Dec, trimester 3 & none; context: trimester 3 & none & missed alarm & no symptom for being past the due date $\rightarrow$ add an overdue-pregnancy symptom and rule & Clinicians \\
\addlinespace
In the 5th month of pregnancy, if there is swelling on the baby's lungs, what should be done? & HRPW, no due date & breathing difficulty; context: trimester 2 & Breathlessness $\rightarrow$ emergency & false alarm & hypothetical question about the fetus read as a symptom of the mother $\rightarrow$ extract only symptoms the patient reports & ML team \\
\addlinespace
It's been 1 month since my C-section, bleeding still not stopped & ANC, no delivery date & prolonged postpartum bleeding; context: PNC & Postpartum bleeding $\rightarrow$ emergency & false alarm & rule fires on any prolonged bleeding $\rightarrow$ narrow the rule & Clinicians \\
\bottomrule
\end{tabular}
\caption{Analysis of the mistakes by \sys{} on the validation set. For each query, it is easy to isolate the mistake to the respective component and assign clear owners with actionable next steps.}
\label{tab:trace-vs-reasoning}
\end{table*}

\subsection{Context inference shows no measurable gain}

\sys{}-NoCtx and \sys{} both reach a recall of 0.810, and their F1 intervals almost entirely overlap (0.706 [0.659, 0.752] against 0.702 [0.655, 0.748]). So, we do not see any performance gain from additionally inferring patient context. \sys{}-NoCtx is also the fastest with a median latency of 1.96 seconds per query. However, it remains necessary for auditing rules whose applicability depends on the patient context.

\subsection{Auditability comes at a cost}

\sys{} is 1.5 times costlier than LLM-Rules. At our current stage, reducing missed emergencies and over-escalations in production simultaneously is our highest priority. Auditability is key to enabling this as outlined in Section~\ref{sec:extraction-audit}. So, we are willing to bear the cost penalty for now while recognizing that it is a genuine obstacle as we scale.

\subsection{The pattern holds across languages}

Table~\ref{tab:by_language} compares LLM-Rules and \sys{} by language. The subsets are small, from 40 to 127 queries, and the confidence intervals overlap. None of the performance differences are statistically significant. Nevertheless, LLM-Rules has the higher recall in four languages while \sys{} has higher precision in five, the same pattern as Table~\ref{tab:architecture}.

\subsection{Performance varies across care categories}

Table~\ref{tab:by_category} compares the two systems on the three care categories with more than 40 queries. The intervals overlap again, so none of the differences are significant. But the pattern is different from Table~\ref{tab:by_language}. For high-risk pregnancy cases, \sys{} leads on both precision and recall. This is the group with the highest risk. In routine antenatal care (ANC), LLM-Rules has the higher recall. For postnatal care queries (PNC), almost half of the escalations are incorrect.

\subsection{Alternatives we discarded}
\label{sec:llmllm}

We did not fine-tune a model on the task, even though we have the labelled data to try it because that would move the clinical policy into model weights that the medical team can neither read nor edit. Every rule revision would then require retraining a new model, significantly slowing down error resolution.

\section{Auditability in deployment}
\label{sec:extraction-audit}

\paragraph{Every error is traceable to a step.} The decomposition maps every mistake to a mistranslation, a symptom either not extracted or missing from the vocabulary altogether, the wrong patient context inferred, or a rule that was wrong or missing. Table~\ref{tab:trace-vs-reasoning} shows four examples of mistakes made by \sys{} on the validation set, two missed alarms and two false alarms. The emergency described by the first query in the table was missed even though the correct symptom was extracted because the relevant rules were valid only for a patient enrolled under the HRPW (high-risk pregnancy) care category, whereas the patient asking the query was enrolled under GHW (general health and well-being). The solution is for the clinicians to discuss the case and make the rule apply more broadly. Similarly, the third query was incorrectly flagged as an emergency because the LLM extracted "breathing difficulty" from a hypothetical question about the unborn baby, even though the mother reported no symptom of her own. The vocabulary and rules are correct here. The ML team needs to ensure the LLM does not hallucinate and extracts only symptoms the patient actually reports.

\paragraph{Each mistake has a clear owner.} Errors in the symptom vocabulary and rule set are handled by clinicians, whereas extraction errors by the LLM are analysed jointly with the ML team. The assignment of clear owners reduces the time to act on a mistake. For example, a clinician can add a new rule without requiring the ML team to run an evaluation as the rule engine operates independently of the LLM.

\paragraph{Evidence from deployment.} \sys{} was deployed to production in June 2026. Since then, it has triaged 152,421 patient queries and flagged 28,535 (18.7\%) as emergencies. Nurses marked 17.8\% of flagged cases as over-escalations and did not report more missed emergencies than before, while the clinicians have added 48 new rules on their own so far.

\section{Conclusion}

We presented the journey of how our emergency triage system evolved after deployment. It began as an opaque, end-to-end LLM pipeline that read a flat list of danger signs, unlike how clinicians work. Given the scale of queries we receive, the number of false alarms soon became a bottleneck, since the system was optimized to reduce missed alarms. Instead of optimising a specific solution to address the current issues, we focused on fixing the process for identifying mistakes and resolving them. Decomposing our system into multiple steps let us trace each mistake to its source, which enabled clinicians to take more ownership of the entire process and resolve issues independently, whereas earlier any change required coordination between multiple teams, resulting in high error resolution times. 

A broader learning is that for domain-specific solutions, it is more important to mimic how domain experts operate than to optimise the model that powers it. Equally important is setting up the right process: one that empowers the domain experts to own the quality of the solution in production, lets them audit issues and trace mistakes to specific components with clear owners, and removes any unnecessary friction in resolving them.

\section{Future Work}

We are expanding our dataset to increase the sample size and coverage across danger signs, languages, and patient contexts, involving more clinicians in the review process. We are also setting up a process to check the clinical accuracy of nurse overrides, to assess their validity as a channel for continuous collection of ground-truth labels beyond a noisy signal of quality in production.

\section{Limitations}
\label{sec:lessons}

\paragraph{Dataset biases.} The metrics have meaningful variance on the current dataset. The labels follow our internal protocols, designed for the families we serve. So the results may not generalize to other settings. 

\paragraph{System limitations.} An emergency whose symptoms are missing from the vocabulary map, or a clinical scenario not yet covered by the rules, will not be caught. Many rules depend on the right patient context. If the query does not specify it, and the stored patient profile either lacks it or holds an outdated value, the emergency classification will be wrong. Nurses are trained to probe patients for missing context, but our system currently does not do it.

\paragraph{Clinical outcomes.} We do not yet have long-term validation of how this emergency triage affects patient outcomes. That will be measured through an upcoming randomized controlled trial evaluating the overall impact of our WhatsApp chatbot.

\section*{Ethical Statement}

Our system is a decision-support tool for our nurses, not a patient-facing or autonomous system. It does not communicate with patients directly or give medical advice. Nurses and in-house doctors review escalations and hold final clinical authority. A missed escalation can cause serious harm, while over-escalation raises patient anxiety and the burden on an already overburdened public health system. Balancing precision and recall is a clinical judgement with real impact on human lives, not a purely technical one. All annotators were compensated fairly. No patient data is used to train any model.

\section*{Acknowledgments}

We would like to acknowledge the contributions of the larger Noora Health team that made this work possible. We also acknowledge hospitals, nurses, and other government stakeholders who supported this study. Finally, we would like to acknowledge patients and families who trusted us with their questions.

\bibliography{references}

\end{document}